\documentclass{article}

\usepackage{microtype}
\usepackage{graphicx}
\usepackage{subfigure}
\usepackage{enumitem}
\usepackage{booktabs} 

\usepackage{hyperref}

\usepackage[accepted]{icml2024}

\usepackage{amsmath}
\usepackage{amssymb}
\usepackage{mathtools}
\usepackage{amsthm}
\usepackage{listings}

\usepackage[capitalize,noabbrev]{cleveref}

\theoremstyle{plain}

\theoremstyle{definition}

\theoremstyle{remark}

\usepackage[textsize=tiny]{todonotes}

\icmltitlerunning{The Landscape and Challenges of HPC Research and LLMs}

\begin{document}
\pagestyle{plain}

\providecommand{\chen}[1]{{\protect\color{purple}{\bf [Le: #1]}}}
\providecommand{\plantodo}[1]{{\protect\color{red}{\bf [contributor: #1]}}}
\providecommand{\chentodo}[1]{{\protect\color{green}{\bf [Todo: #1]}}}
\providecommand{\nka}[1]{{\protect\color{violet}{\bf [Nesreen: #1]}}}

\twocolumn[
\icmltitle{The Landscape and Challenges of HPC Research and LLMs}
           




\begin{icmlauthorlist}
\icmlauthor{Le Chen}{isu}
\icmlauthor{Nesreen K. Ahmed}{intel}
\icmlauthor{Akash Dutta}{isu}
\icmlauthor{Arijit Bhattacharjee}{isu}
\icmlauthor{Sixing Yu}{isu}
\icmlauthor{Quazi Ishtiaque Mahmud}{isu}
\icmlauthor{Waqwoya Abebe}{isu}
\icmlauthor{Hung Phan}{isu}
\icmlauthor{Aishwarya Sarkar}{isu}
\icmlauthor{Branden Butler}{isu}
\icmlauthor{Niranjan Hasabnis}{intel}
\icmlauthor{Gal Oren}{Technion}
\icmlauthor{Vy A. Vo}{intel}
\icmlauthor{Juan Pablo Munoz}{intel}
\icmlauthor{Theodore L. Willke}{intel}
\icmlauthor{Tim Mattson}{human}
\icmlauthor{Ali Jannesari}{isu}

\end{icmlauthorlist}

\icmlaffiliation{isu}{Department of Computer Science, Iowa State University, Ames, USA}
\icmlaffiliation{intel}{Intel Labs, USA}
\icmlaffiliation{Technion}{Technion - Israel Institute of Technology, Isreal} 
\icmlaffiliation{human}{Human Learning Group, USA}
\icmlcorrespondingauthor{Le Chen}{lechen@iastate.edu}

\icmlkeywords{Machine Learning, ICML}

\vskip 0.3in]



\printAffiliationsAndNotice{}  

\begin{abstract}
Recently, language models (LMs), especially large language models (LLMs), have revolutionized the field of deep learning.
Both encoder-decoder models and prompt-based techniques have shown immense potential for natural language processing and code-based tasks.
Over the past several years, many research labs and institutions have invested heavily in high-performance computing, approaching or breaching exascale performance levels. 
In this paper, we posit that adapting and utilizing such language model-based techniques for tasks in high-performance computing (HPC) would be very beneficial.
This study presents our reasoning behind the aforementioned position and highlights how existing ideas can be improved and adapted for HPC tasks.

\end{abstract}

\section{Introduction}
Large Language Models (LLMs), such as GPT-4 ~\citep{OpenAI_GPT4_2023}, represent the forefront of artificial intelligence, specifically in understanding and processing human language. 
Predominantly built on the transformer architecture ~\citep{vaswani2017attention}, these models are meticulously trained on extensive textual datasets that encompass a diverse range of human discourse.
Characterized by their hundreds of millions (or even billions) of trainable parameters, LLMs have demonstrated strong capacities in interpreting complex human language and solving a myriad of Natural Language Processing (NLP) tasks. 

Following the state-of-the-art results achieved by LLMs in NLP tasks, there is growing research interest in applying LLMs to Programming Languages (PL) tasks. 
Recent work ~\citep{chen2023lm4hpc, chen2023data, kadosh2023scope, ding2023hpc} have explored the use of LLMs for code and programming tasks (e.g., code generation, translation, completion, etc), yielding impressive results.
This highlights the potential of applying LLM to other domains, such as high-performance computing (HPC). 

HPC is a specialized domain that utilizes parallel processing techniques on modern multi/many-core architectures to solve large-scale complex computational problems.
HPC is instrumental in a wide range of critical applications, from climate modeling, computational chemistry, and biomedical research to astrophysical simulations. HPC provides a framework for scalable processing and analysis of complex problems with massive datasets, which makes it a cornerstone in advancing scientific and technological frontiers. Therefore, the application of LLMs to HPC is garnering increasing interest. 
Recent pioneering works~\citep{chen2023lm4hpc, nichols2023modeling, chen2023data} have begun to explore the application of LLMs to solve tasks within the HPC domain such as parallel code generation. 
These works indicate a promising synergy between large language models and HPC.


Despite the recent success of LLMs in programming language-related tasks such as code generation and translation, there are no detailed studies of applying LLMs to HPC tasks. 
We conjecture that the unique characteristics of HPC pose distinct challenges in integrating LLM and HPC. 
To date, there has been no comprehensive exploration of the full potential or challenges of adapting and utilizing LLMs within the HPC domain. 
This represents a notable gap in our understanding, especially considering the rapid advancements in both fields. 
In this work, we aim to address this gap by systematically examining the reciprocal benefits that can arise from the intersection of LLMs and HPC. Further, we illustrate through a case study presented in Section~\ref{sec:case} how tailored designs can facilitate mutual advantages between LLMs and HPC. 
This paper investigates ways in which LLMs can be tailored and optimized for HPC tasks while also addressing the challenges that accompany such integration. 


\section{Background}
\label{sec:background}

 \begin{figure*}[ht]  
\centering 
\includegraphics[width=0.8\textwidth]{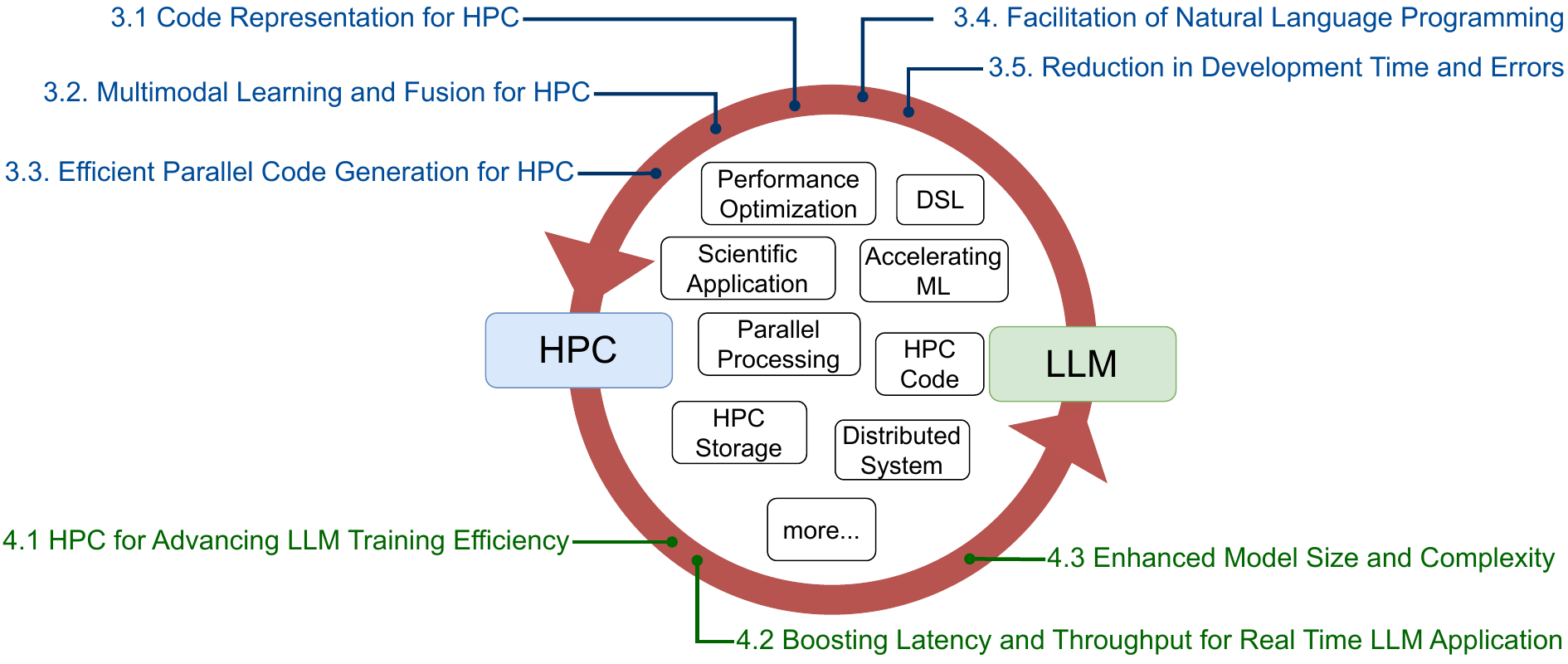} 
\caption{
A visual abstract of our paper highlights critical areas of exchange between the fields of high-performance computing (HPC) and large language model (LLM) research (outer labels). We describe several downstream HPC applications to target (center, black outline boxes). Focusing on these areas and applications will create a virtuous cycle of improvement that advances both fields.
} 
\label{fig:pipeline}
\end{figure*}

\textbf{Large Language Models (LLMs)}, especially those based on the Transformer architecture like GPT-4, represent a major leap in Artificial Intelligence (AI) and Natural Language Processing (NLP). These models, trained on vast textual datasets, excel in understanding and generating human language, demonstrating capabilities across a range of NLP applications.

\textbf{LLM for Programming Language-Related Tasks}
LLMs like GPT4 and CodeLlama~\citep{roziere2023code} have been trained with both human language and programming language corpus. The knowledge of code has led to remarkable outcomes of these models in various programming language tasks, such as code generation~\citep{poldrack2023ai, achiam2023gpt}, code explanation~\citep{khan2022automatic}, and software testing~\citep{schafer2023empirical}.  These achievements underscore the models' capabilities in supporting a wide range of programming languages, enhancing code completion accuracy, and facilitating interactions between natural language (NL) and programming language (PL). They also have inspired recent attempts~\citep{chen2023lm4hpc, nichols2023modeling, chen2023data} to apply LLMs in HPC tasks. However, the high-performance computing (HPC) domain presents distinct interests and requirements when integrating LLMs.


\noindent
\textbf{High-Performance Computing (HPC)}, characterized by its ability to rapidly process and analyze large datasets, is pivotal in fields ranging from scientific research to data analytics. Its evolution has seen significant advancements in computational power and efficiency, making it indispensable for complex problem-solving tasks. 
An \textbf{HPC ecosystem} refers to the comprehensive environment that encompasses all the hardware, software, workflows, networking, and storage solutions designed to support and enhance high-performance computing tasks. 

\textbf{HPC Tasks}, as defined in this paper, encompass challenges and problems addressed within the HPC ecosystem. For a detailed list of HPC tasks that have been investigated using Machine Learning (ML) techniques, please refer to Appendix~\ref{sec:appendix-tasks}. The uniqueness of HPC tasks stems from the distinct characteristics of the HPC field itself. For instance, consider the task of code generation within HPC: this specific task emphasizes the generation of parallel code and ensuring compatibility with parallel computing frameworks such as OpenMP and MPI, highlighting the specialized focus required for HPC-related endeavors.

\textbf{Challenges.} Integrating LLMs with HPC presents an exciting frontier research but also challenges. Bridging the gap between the advanced capabilities of LLMs and the specialized requirements of HPC tasks involves overcoming several hurdles. Below, we outline some common challenges that necessitate innovative approaches and solutions.

\begin{itemize}[topsep=0pt, partopsep=0pt, itemsep=0pt, parsep=0pt]
    \item HPC domain-specific knowledge: The effective application of LLMs in HPC requires a deep understanding of domain-specific languages (DSLs) and HPC-related code information, including performance metrics and Memory I/O details. 
    \item Methodology to Leverage HPC Tools: The strategies for LLMs to utilize the plethora of tools within the HPC ecosystem remain underexplored. Developing methodologies that can seamlessly integrate these tools with LLM capabilities is crucial for enhancing the performance and efficiency of HPC tasks.
    \item Data Representation and Integration: Effectively representing HPC task data in a form that is usable for LLMs is a non-trivial challenge. This includes converting complex scientific data and parallel code structures into formats that LLMs can process and learn from.
    \item Performance Metrics and Evaluation: Developing appropriate metrics to evaluate the performance of LLMs in the context of HPC tasks is crucial. Traditional NLP metrics may not adequately capture the effectiveness of LLMs in generating or optimizing HPC code, necessitating the development of specialized benchmarks and evaluation frameworks.
\end{itemize}

\section{LLMs for HPC: Pathways and Directions}
\label{sec:llm4hpc}

This section delves into the emerging frontiers of integrating Large Language Models within the High-Performance Computing HPC domain. Initially, it delineates various potential pathways and strategic directions to apply LLMs in HPC. Subsequently, the discussion shifts to examining the state-of-the-art in code-based LLMs, showcasing the latest advancements and innovations in this area.
\subsection{Code Representation for HPC}
\textbf{Code representation} refers to the way source code is modeled to facilitate understanding, analysis, or processing by humans and machines. The most intuitive way to represent code is using the source code text. Previous works have utilized this textual representation of code with LLMs for various code tasks such as code explanation~\citep{sarsa2022automatic, macneil2023experiences, macneil2022generating}, code completion~\citep{chen2021evaluating, li2023starcoder, roziere2023code}, and code similarity checking~\citep{saieva2023contrastive, jiang2023nova}. However, in the realm of HPC, there is a pronounced emphasis on performance optimization. This necessitates the exploration of alternate code representations to work with LLMs for the unique demands and efficiencies required for HPC.


Modern HPC code is predominantly written in compilable languages such as C, C++, and Fortran, which allows fine-grained control over system resources.
Compilers, like LLVM, provide a lower-level representation of such compilable source code called Intermediate Representation (IR).
LLVM IR helps bridge the gap between high-level source and low-level machine code, and most compiler-driven performance optimizations are applied on the IR level. 
Previous research efforts have leveraged LLVM IR in conjunction with machine learning techniques to analyze and model semantic and structural dependencies in source code, significantly enhancing performance in HPC tasks.
For example, PERFOGRAPH ~\citep{tehranijamsaz2023perfograph} and PROGRAML ~\citep{cummins2021programl} both used IR-level representation with graph neural networks for HPC tasks such as parallelization detection, device mapping, and non-uniform memory access (NUMA) configuration prediction. 
The quality of their results compared to source textual representations indicates that for HPC tasks, such structural and semantic features exposed through IRs are very important.
Additionally, the work of \citet{mahmud2023autoparllm}, which is among the first to employ IR with LLMs for generating parallel code, further reinforces this viewpoint. 
Their evaluation illustrates that LLMs adapted as HPC code representations can effectively tackle HPC tasks.

\textbf{Challenge}. Creating various representations of code is not always straightforward. It demands an in-depth understanding of both the programming language in use and the array of tools available within the HPC ecosystem. Moreover, there is no assurance of successful conversion for all code data. For instance, the process of converting source code into LLVM IR requires compilation, which means that code containing errors might not generate its corresponding IR. Consequently, the code dataset might not fulfill the requirements for Large Language Models (LLMs).  

The HPC ecosystem offers promising strategies to overcome this hurdle. A widely adopted method to augment the dataset size involves employing various code mutation techniques \citep{chen2022multi, chen2023learning, tehranijamsaz2023perfograph, mahmud2023autoparllm}. Tools such as SymPy~\citep{meurer2017sympy} and Jinja2~\citep{ronacher2008jinja2} have been utilized in previous code mutation approaches~\citep{chen2022multi, chen2023learning}.  Additionally, a different approach to expanding the dataset involves manipulating the compilation flags of LLVM-IR, facilitating transformations at the IR level \citep{tehranijamsaz2022learning, chen2022multi, chen2023learning}. This technique allows for the creation of varied IR versions of the same program by applying different combinations of compilation flags.

Additionally, LLMs need to have inherent knowledge of how HPC code executes. For instant, how threads/cores communicate with each other, or what variables need to be shared or private among threads. Without this knowledge, concurrency bugs can easily be introduced into the code generated by LLM. Therefore, a program representation that can expose this inherent knowledge to the LLMs is necessary.

\subsection{Multimodal Learning and Fusion for HPC}
\textbf{Multimodal learning} has recently gained significant attention in foundational model (FM) research. 
One primary approach involves encoding data from diverse modalities into a unified latent space, while another one focuses on generating data across various modalities using transformer decoders. 
Pioneering models such as CLIP \cite{radford2021learning}, ALBEF \cite{li2021align}, and ALIGN \cite{jia2021scaling} have introduced innovative cross-modal alignment techniques, demonstrating remarkable capabilities in correlating text and image data. 

We believe that such multimodal language model encoders can be effectively applied to HPC performance optimization tasks. 
These are well-suited for representational learning, where it's crucial to represent both source code and system-related features (like performance counters) within an embedding space. 
Prior research \cite{dutta2023performance} has already illustrated the benefits of multimodal modeling for combining IR-based code features with system runtime features. 
However, these benefits are largely confined to specific systems and tasks. 
An LLM-style multimodal encoder trained on diverse HPC data could significantly enhance performance optimization, offering increased adaptability across various systems and tasks.

Moreover, leveraging the concept of encoder-diffusion, multimodal code generators can potentially emulate the success of image generators like DALL-E \cite{ramesh2021zero}. 
Such models, upon receiving descriptions in natural language, are expected to generate semantically and structurally sound parallel code. 
In the domain of \texttt{pragma}-based parallel programming, these generators could also aid with accurately selecting directives, options, and other runtime factors, thereby augmenting HPC code performance. 
Critical performance bottlenecks, such as data placement/transfer across devices and NUMA domains, stand to benefit immensely from an enhanced code generator that focuses on both performance and correctness. 

\textbf{Challenge.}
Multiple factors impact the performance of HPC code.
Starting from compiler optimizations to system environment, everything plays an important role in the performance of an HPC application.
Therefore, it is necessary to identify and represent most of these factors when dealing with HPC setups.
A key challenge is to use such varied features within the same context.
Multimodality can help in such cases; however, we must be careful about how we encode and fuse each of these varied modalities.
For example, when we try to generate the best performing \texttt{pragmas} for existing serial code, it might be necessary to consider the impact of compiler optimizations (modeled through IRs) and model the impact of system environment through performance counters.

\subsection{Parallel Code Generation using LLMs}
The widespread adoption of multi/many-core processing has increased parallelization efforts to improve code execution efficiency.
We believe LLMs can play a vital role in such parallelization efforts.
In spite of their ground-breaking achievements in code generation, LLMs were shown to struggle to generate high-quality parallel code even in simple cases~\citep{mahmud2023autoparllm}.
~\citep{valero2023comparing} compared the autonomous parallelization of HPC code using Llama-2 and GPT-3 across various programming models and languages, including OpenMP, OpenACC, and CUDA. 
The authors concluded that although LLMs can successfully generate serial codes, the generated parallel code often had issues with correctness and poor speedup. 
Works such as ~\cite{godoy2023evaluation} have also demonstrated that the performance of LLMs decreases with increasing complexity and change among different parallel execution models (e.g., OpenMP vs MPI). This evidence limits the practical application of LLMs in HPC. 

LLMs targeted to the HPC domain might be used to target a variety of generation tasks such as full program/function generation, code completion, optimization suggestions, and code performance predictions.
So far, only a handful of works~\citep{mahmud2023autoparllm,nichols2023modeling, kadosh2023domain, nichols2024can} have attempted to address these open research problems. Other works ~\citep{nichols2023modeling, kadosh2023domain, nichols2024can} have attempted to parallelize sequential programs by inserting directives and clauses of parallel programming models such as OpenMP, OpenACC, etc.
These studies, however, are limited to a small set of parallel configurations, and further research is needed to improve the parallel code generation capabilities of LLMs.

Limited experiments so far~\cite{nichols2023modeling}, have shown the capabilities of LLMs in identifying the better performing code given two options.
However, in reality, compilable programs, which form the backbone of most HPC tasks, can be compiled and executed with many options.
Having an LLM that can evaluate thousands of parameters/configurations prior to execution/compilation (e.g., phase ordering) and predict the most profitable version of a program can revolutionize existing performance modeling techniques, reducing optimization overheads manifold.  

\textbf{Challenge}. One of the major challenges of parallel code generation using LLMs is the lack of existing datasets. 
There have been some attempts to develop parallel code datasets, but as mentioned earlier, most of these attempts only considered a small subset of possible parallel configurations. 
Another major challenge is that the generated codes should i) be correct, and ii) have good performance. 
The work in~\citep{mahmud2023autoparllm} attempts to address this by demonstrating that LLM-generated code can be both correct and perform well. 
But again, the study is based on a limited scope of parallelization techniques. 
Additionally, there is a lack of proper metrics to evaluate the quality of the generated parallel programs. 
BabelTower ~\citep{wen2022babeltower} tries to address this issue, but only for generated CUDA programs. 
Hence, a more extensive study is required to come up with a general set of metrics to evaluate the quality of all generated parallel programs.

\subsection{Facilitation of Natural Language Programming}

\textbf{Natural Language Programming} bridges the human language and programming language, enabling more intuitive interactions with computational systems. The implementation of NLP has primarily been pursued through two key approaches: natural language (NL) to programming language (PL) search and natural language to programming language generation. NL to PL search leverages information retrieval methods, embedding NL queries and code snippets as vectors to facilitate accurate and efficient code searches. This approach is instrumental in deriving the correct code from a vast pool of possibilities. On the other hand, PL generation focuses on creating new code snippets for specific tasks, particularly those not covered in existing training datasets.

In the context of HPC, the integration with LLMs opens up innovative avenues for natural language programming. Within this scope, both NL and PL are re-envisioned to cater to the unique needs of HPC. NL should encompass HPC-specific information, such as optimization requirements and hardware configurations, while PL needs to align with the distinctive characteristics of HPC code, as outlined in Section 3.1. These developments open up novel research directions in NLP for HPC. Firstly, NL to PL tasks in this domain should concentrate on generating HPC-oriented code, marking a shift from the general sequential code generation seen in previous works like CodeBERT \cite{040_CodeBERT} and CodeT5+ \cite{084_CodeT5Plus}. Secondly, the translation from PL to NL should integrate and convey HPC-specific knowledge, ensuring that the generated natural language explanations and instructions are rich in HPC context and insights."



\subsection{Reduction in Development Time and Errors}
Code LLMs have improved the development process of software packages. In particular, they can assist developers in various ways, such as code completion, code understanding, documentation, natural language to code generation, cross-language support, error detection, etc. CoPilot~\cite{copilot} is one of the most well-known Code LLMs and has been integrated into various Integrated Development Environments (IDEs) such as Visual Studio and JetBrain IDEs. Other CodeLLMs include Codeium~\cite{codeium}, Tabnine~\cite{tabnine}, and various other open source code LLMs available on HuggingFace library such as StarCoder~\cite{li2023starcoder}, CodeLLama~\cite{roziere2023code}.

A recent work~\cite{cummins2023large} has shown how Code LLMs can help to find a list of best compiler options to optimize the LLVM assembly for code size. In a similar way, Code LLMs should also be able to help in many other optimization tasks, such as phase ordering, vectorization, etc. 
Moreover, integration of HPC toolkits such as compilers and profilers with LLMs can further improve the performance of Code LLMs for HPC-related tasks. For instance, compiler or profiler feedback (for example, performance counters) can provide additional insight and features to the Code LLMs to solve the downstream task.

However, currently, Code LLMs often struggle when they are faced with HPC-related tasks~\cite{godoy2023evaluation, nichols2024can}. 
Recently, a few works have tried to address this issue \cite{nichols2023modeling, kadosh2023domain}. 
Nonetheless, we believe Code LLMs still have much to gain from HPC technologies while more deeply impacting the coding of HPC applications.

\subsection{State-of-the-art in Code LLMs}
Currently, most code-based LLMs are trained on a multitude of programming languages and natural languages. CodeLlama \cite{roziere2023code}, StarCoder \cite{li2023starcoder}, WizardCoder \cite{luo2023wizardcoder}, PolyCoder \cite{xu2022systematic} are some notable examples of such. CodeLlama is a code-specialized version of Llama 2 that was trained with code-specific datasets.CodeLlama is available in variety of parameter sizes - 7B, 13B, 34B and 70B. Each trained with 500B tokens of code and code-related related data except for 70B which is trained with 1T tokens. It supports most of the common programming languages like Python, C++ and Java. StarCoder is a LLM trained on permissively licensed data from GitHub with 80+ programming languages. The model is specifically trained on the Stack \cite{kocetkov2022stack} dataset. Its a 15B model trained for 1T tokens. WizardCoder builds on top of StarCoder with its application of Code Evol-Instruct. It posted surprisingly good HumanEval benchmark results, only falling behind GPT-3.5 and GPT-4 which are much larger in parameter count. PolyCoder is another LLM trained using a GPT-2 based model with parameter size of 2.7B. It was trained on 249GB of code data across 12 programming languages.\\
All these models above are trained on a multitude of programming languages. However, the HPC community focuses on only three core programming languages, C, C++, and Fortran. Training a domain-specific model specifically for HPC could help us significantly reduce the parameter size and reduce the time taken for training and inference. This in turn will also benefit from using a smaller amount of compute resources. CompCoder \cite{kadosh2023scope} is an LM that was specifically trained with HPC code. Another state-of-the-art is HPC-GPT \cite{ding2023hpc} which is a Llama-based model specifically tuned for HPC programming. \cite{chen2024ompgpt} talks about an LLM which is designed specifically for OpenMP pragma generation which is an important task in parallelism.


\section{Advantages of Integrating LLMs with HPC}
\label{sec:hpc4llm}

To this end, we have explored various pathways and directions for applying LLMs within the realm of HPC. Conversely, harnessing the power of HPC to enhance LLM capabilities presents a wealth of potential applications. This symbiotic relationship between HPC and LLMs opens the door to groundbreaking advancements in computational efficiency and intelligent processing.

\subsection{HPC for Advancing LLM Training Efficiency}

The \textbf{efficient training} of Large Language Models such as GPT-3~\cite{john2023opengpt} and LLaMA \cite{touvron2023llama} 
has become a cornerstone in the field of natural language processing. These models have exhibited outstanding capabilities in various natural language understanding tasks
\cite{brown2020language, devlin2018bert}. However, a critical bottleneck in the evolution of these models is their training inefficiency, often marked by extended time durations and the need for vast computational resources, primarily GPUs.

Integrating HPC in training LLMs is beneficial and essential for the progression of natural language processing. 
This position is supported by emerging research, notably by \citet{smith2022using} and \citet{Zhang2022}, which illustrated the profound impact of HPC in reducing training times through advanced parallel computing and network architectures.
The traditional approach to LLM training is increasingly untenable due 
to these models' growing complexity and size. 
HPC offers a scalable solution to this challenge, enabling more efficient utilization of resources and drastically reducing training time. This enhancement is crucial for the feasibility of creating sophisticated models and 
ensuring that advancements in this field are sustainable and accessible. Adopting HPC in LLM training has far-reaching implications. It paves the way for more rapid development cycles, enabling researchers and developers to iterate and innovate at an accelerated pace. Furthermore, it democratizes access to advanced model training, potentially broadening the scope of research and application in both academic and industrial settings. In conclusion, integrating HPC in the training of LLMs is a pivotal step forward for artificial intelligence. It addresses current limitations and sets a foundation for more groundbreaking developments in LLM. We advocate for increased investment and research in the transformative potential of HPC in the realm of LLM training.

Another crucial HPC contribution relates to improving the data preprocessing throughput. 
LLMs necessitate a foundational pre-training phase on extensive corpora, encompassing large, organized sets of machine-readable texts. This process requires substantial data preprocessing, encompassing critical tasks such as tokenization, sentence segmentation, and noise removal~\citep{conneau2019cross}. Ensuring the quality of the collected text through preprocessing is crucial since it can profoundly influence the capability and performance of LLMs, as highlighted by recent research~\citep{moore2010intelligent, kaddour2023challenges, chen2023compcodevet}.
HPC has a rich history of supporting distributed algorithms over many years. Recent works~\cite{luengo2020big} have harnessed HPC solutions for data preprocessing, capitalizing on its scalability and efficiency advancements. These advantages translate into tangible benefits, such as significant time reduction, the empowerment of real-time applications, and the assurance of cost-effectiveness and environmental sustainability in the preprocessing phase. Leveraging HPC solutions for Large Language Model (LLM) data preprocessing represents a promising approach, 
effectively addressing the computational demands inherent in this critical preprocessing step.


\subsection{
Boosting Latency and Throughput for Real-time LLM Applications }

HPC is critical in mitigating \textbf{latency for real-time LLM applications}, which are inherently resource-intensive and challenging to deploy at the user end due to their computational demands. The autoregressive nature of LLMs, which predicts each word based on the preceding context, necessitates significant memory and computational resources, limiting parallel task processing and speed.
Cloud-based LLM deployments face latency issues, particularly under heavy query loads due to the sequential generation of tokens. This process, where the response time increases with the context length, underscores the inefficiency of conventional cloud services for real-time LLM applications. HPC architectures, characterized by their thousands of heterogeneous nodes, provide a scalable solution by enabling parallel processing of LLM tasks, significantly reducing response times. Intel's Data Center GPU Max Series exemplifies the hardware optimization for such applications, offering high-density parallel computing capabilities essential for AI and HPC workloads.
Furthermore, addressing the unpredictable prompt lengths in user queries requires dynamic memory and processing strategy adjustments, highlighting the complexity of optimizing LLM inference for consistent performance and efficiency. The integration of HPC with advanced computational models and hardware innovations thus emerges as a key strategy for reducing latency in LLM applications, facilitating their deployment in scenarios demanding quick response times.

Inference of LLMs is an inherently sequential process, making full utilization of many accelerators a complex problem. Compounding this, many LLMs 
need to fit 
within the memory of a single accelerator, necessitating the usage of many accelerators \cite{pope2022efficiently}. Novel parallelization techniques such as Megatron-LM \cite{shoeybi2020megatronlm} or DistributedLlama \cite{tadych2024distributedllama} increase utilization via tensor parallelism. While tensor parallel techniques allow many accelerators to contribute to the computation of individual tensor operations, significant communication is required between the accelerators. Such high communication requirements can result in extreme interconnect bandwidth
bottlenecks, even for extremely slow compute nodes like Raspberry Pi 4s. Even as few as eight Raspberry Pi nodes become bottlenecked by Gigabit Ethernet \cite{tadych2024distributedllama}. HPC uniquely suits inference strategies with high communication requirements, commonly employing multiple accelerators in a single node and high bandwidth communication between nodes \cite{10.1145/3581784.3613215}. HPC clusters can also allocate substantial resources in unique and efficient architectures, revealing novel strategies for 
optimizing 
latency or throughput \cite{pope2022efficiently}.

\subsection{Enhanced Model Size and Complexity}
The architecture of LLMs is the fundamental aspect that contributes to their \textbf{size} and \textbf{complexity}. Recent works in LLMs have seen a consistent increase in the size of the models, with the state-of-the-art models boasting billions and trillions of parameters such as Llama-2 \cite{touvron2023llama}, BLOOM \cite{workshop2022bloom}, Megatron-Turing NLG \cite{smith2022using}, and GPT-3 \cite{brown2020language}. The increase in size is not merely a quantitative change but brings about a qualitative transformation in the model's performance. Larger models have demonstrated remarkable improvements in various NLP tasks \cite{koroteev2021bert, sallam2023utility, roumeliotis2024llms}. The depth and breadth of these tasks correlate to the models' size and complexity, and the interplay between them drives the effectiveness of LLMs.

However, the most immediate impact of larger LLMs is the significant increase in computational power required for their training and inference. Models with trillions of parameters require huge processing power and memory, pushing the boundaries of current computing capabilities \cite{smith2022using}. HPC systems, with their extraordinary scale, are uniquely equipped to meet these demands \cite{10.1145/3581784.3613215}.  This intersection is not one-sided; LLMs can also significantly contribute to the evolution of HPC as they serve as demanding applications that drive the development of more sophisticated HPC architectures and software. This includes advanced machine learning frameworks, optimization algorithms, and parallel computing techniques, which are essential for efficiently managing the computational demands of LLMs \cite{john2023opengpt}.

Future LLMs will likely continue to grow in size and complexity, potentially increasing by several orders of magnitude. Therefore, we believe that this growth can fuel the advancements in HPC and finally bridge the gap between AI and human cognition more closely.

\section{Ethical Considerations}
\label{sec:Ethical}
HPC has been at the forefront of scientific and technological advancement. With LLMs entering this realm, ethical questions arise. Concerns include data privacy, consent, and responsible AI deployment. It is vital to uphold ethical standards and ensure that LLMs in HPC adhere to established guidelines and regulations. 
LLMs are trained on extensive datasets, potentially mirroring societal biases. When integrated into HPC applications, this bias can inadvertently influence results. Detecting and mitigating bias in LLM-generated content is imperative to prevent skewed outcomes in scientific research or decision-making processes.
Transparency in LLM-HPC integration is crucial. Users and stakeholders must understand how LLMs impact computations and results. Establishing accountability mechanisms for LLM-generated content within HPC applications helps maintain trust and integrity.
Addressing ethical concerns and bias in HPC with LLMs necessitates proactive measures. Strategies may include rigorous data curation, bias-aware training, ongoing monitoring, and the development of fairness metrics specific to HPC use cases.
Implementing ethical AI governance frameworks ensures responsible LLM deployment. Collaborative efforts between AI practitioners, ethicists, and domain experts are vital to navigating complex ethical landscapes and establishing ethical guidelines for LLM-HPC integration.


\section{Case Studies and Potential Gaps}
\label{sec:case}
\subsection{Instances of LLMs Enhancing HPC Performance}

In this section, we discuss an application of LLMs to the HPC problem of automatically generating parallel programs for shared memory systems (using OpenMP pragmas).

Shared memory systems are characterized by multiple compute cores (e.g., CPU cores) that share access to common caches (e.g., L3 cache). For instance, systems based on the 5th generation Intel Xeon processor (codenamed Emerald Rapids)~\cite{rapids}, contain anywhere between 8 to 64 cores, all of which share access to the last level cache (L3 typically). Getting the best performance out of such systems requires writing parallel code, which divides the problem into subproblems and executes them in parallel on different cores. Writing a parallel version of serial code, however, is tricky, courtesy of typical multi-threading problems --- it requires reasoning of data dependence, race conditions, deadlocks, etc. Programming standards such as OpenMP simplify this task considerably to the extent that OpenMP is the most popular parallel programming API in open-source \cite{kadosh2023quantifying}. 

\begin{figure}[h]
\centering
\begin{minipage}{.48\textwidth}
\begin{lstlisting}
// Serial code for element-wise multiply
for (int i = 0; i < a.size(); i++) {  
  a[i] = b[i] * c[i];
}
\end{lstlisting}
\end{minipage}
\hfill
\begin{minipage}{.48\textwidth}
\begin{lstlisting}
// Parallel version of the above serial code
#pragma omp parallel for
for (int i = 0; i < a.size(); i++) {  
  a[i] = b[i] * c[i];
}
\end{lstlisting}
\end{minipage}
\caption{Comparison between serial and parallel implementations of element-wise multiplication.}
\label{fig:serial_vs_parallel}
\end{figure}

As an example, the first code snippet in \autoref{fig:serial_vs_parallel} shows a serial version of code that performs element-wise multiplication on two \texttt{std::vector}s, while the following code snippet shows the parallel version of the serial code. The \texttt{\#pragma omp parallel for} pragma causes the OpenMP runtime to create a team of threads, where each thread operates on an individual subset of the iteration space, leading to the better utilization of underlying multiple compute cores. While standard compilers, such as GCC, LLVM, etc., and source-to-source translation tools (S2S), such as Cetus~\cite{dave2009cetus}, AutoPar~\cite{dever2015autopar}, Par4All~\cite{creusillet2009par4all}, ComPar~\cite{mosseri2020compar}, etc., can automatically generate parallel versions of serial code, they, however, had limited success~\citep{harel2020source,prema2017identifying,prema2019study}, especially because of a lack of robustness.

The limitations of the existing tools in automatically generating parallel versions of serial code have led to the introduction of AI-based tools for programming assistance. Instead of relying on formal program analysis passes (such as loop dependence analysis in compilers), AI-based tools for this problem leverage recent advancements in the field of NLP (especially Transformer architecture) to accurately determine the parallelization potential of code. A simple categorization of these AI-based tools could be as follows: (1) \emph{OpenMP-specific tools}, such as PragFormer~\citep{harel2023learning,kadosh2023pragformer}, OMPify~\cite{kadosh2023advising}, Graph2Par~\cite{chen2023learning}, HPCoder~\cite{nichols2023modeling}, AutoParLLM~\cite{mahmud2023autoparllm}, etc., that are solely designed for the OpenMP parallelization problem, (2) \emph{Pre-trained HPC-oriented models} that are the fine-tuned for OpenMP, such as MonoCoder~\cite{kadosh2023domain} and OMP-GPT~\cite{chen2024ompgpt}, and (3) \emph{general-purpose tools}, such as ChatGPT, CodeLLaMa~\cite{roziere2023code}, etc., that can solve the OpenMP parallelization problem, in addition to several other programming related or unrelated tasks~\citep{godoy2023evaluation, valero2023comparing, nichols2024can}. We will review these tools along with different design choices. (Since the last category of tools are not specifically designed for the OpenMP parallelization problem, we will not discuss their design choices.) 


\begin{itemize}[leftmargin=*]

    \item \emph{Problem formulation:} The problem of automatically parallelizing serial code using OpenMP can be divided into multiple subproblems. To be precise, the problem that these approaches attempt to solve can be defined as: \emph{Given a piece of serial code (mostly \texttt{for} loops), determine if the code can be parallelized, and if so, suggest appropriate OpenMP pragma.} As the first part of the problem statement 
    is a boolean question, tools such as PragFormer, OMPify, and Graph2Par formulate it as a binary classification problem (this same formulation also applied to other parallelization strategies, such as MPI (e.g., MPI-rical~\cite{schneider2023mpi}). Once these approaches determine the parallelization potential of a loop, 
    the next subproblem is to suggest appropriate OpenMP pragma as a multi-class classification problem. Specifically, Graph2Par considers four specific items from OpenMP (\texttt{target}, \texttt{simd}, \texttt{private}, \texttt{reduction}) 
    that could apply to a parallel loop. PragFormer and OMPify, on the other hand, consider two additional OpenMP clauses (\texttt{private} and \texttt{reduction}). Given the large number of clauses, library functions, and pragmas in OpenMP~\cite{kadosh2023quantifying}, these approaches have a long way to go before the full range of OpenMP can be applied to HPC programming problems.
    
    \item \emph{Source code representation:} The representation of the input serial code, is an important design decision for this problem as the accurate predictions depend upon the ability of the AI model to learn to reason about certain program properties (such as loop-carried dependence) that determine the parallelism potential. Treating source code as text and employing a sequence of tokens representation did not yield satisfactory results~\cite{kadosh2023advising}, consequently, all of these approaches have leveraged sophisticated compiler-based code representations such as abstract-syntax tree (AST), data-flow graph (DFG) (in OMPify), or even specialized ones such as heterogeneous augmented abstract syntax tree (Augmented-AST) in Graph2Par~\cite{chen2023learning}. Also, some of these approaches have devised new tokenization strategies. For instance, Kadosh et. al. have devised TokomPiler~\cite{kadosh2023scope} to address specific requirements of preprocessing HPC code (written mostly in C, C++, and Fortran) and compilation-centric tasks.
    
    \item \emph{Training dataset:} The lack of curated, publicly-available datasets has forced teams working on these techniques to synthesize their own training datasets using various sources such as open-source programs containing OpenMP pragmas, parallel programming benchmarks (e.g., NAS parallel benchmark~\cite{bailey1991parallel}), etc. Specifically, a common approach followed for synthesis is to search C/C++ programs containing \texttt{for} loops that have OpenMP parallel loops (e.g., \texttt{\#pragma omp parallel for}). The \texttt{for} loops are then used as input to the model, while their OpenMP pragmas (or their lack of) are used to generate appropriate labels. Thankfully, authors of these approaches have released their datasets publicly for further research (e.g., OMP\_Serial by Graph2Par, Open-OMP by PragFormer). The most comprehensive HPC-oriented training dataset to this date is the HPCorpus~\cite{kadosh2023quantifying} dataset, containing a total of 300K repos, 70 GB, 9M files across C, C++, and Fortran code from GitHub, with hundreds of thousands of those functions able to compile successfully~\cite{chen2023compcodevet}. 
    This repo includes common parallel programming APIs, such as MPI, CUDA, OpenCL, TBB, Cilk, OpenACC, and SYCL.
    
    \item \emph{Model architecture:} These approaches employ popular deep learning innovations such as Transformer architecture, graph neural networks (as source code can be represented as a graph), etc., to find parallelism opportunities within serial code and then generate parallel versions by automatically inserting OpenMP pragmas. Specifically, Graph2Par uses a modified transformer model called heterogeneous graph transformer (HGT)~\cite{hu2020heterogeneous}, while OMPify builds on top of GraphCodeBERT\cite{guo2020graphcodebert}, a pre-trained model for programming languages that considers the inherent structure of the code by accepting source code along with its dataflow graph. Models employed by these approaches are typically smaller than LLMs such as CodeLLaMa, GPT-3.5, etc., that can also parallelize serial code. In spite of the smaller sizes, these approaches have outperformed larger models such as ChatGPT on the task of parallelizing serial code~\cite{kadosh2023domain, chen2024ompgpt}.

    \item \emph{Results:} Overall, better and problem-specific code representations have helped these OpenMP-specific approaches outperform code LLMs on the OpenMP parallelization problem. Specifically, PragFormer has shown that it can outperform a formal, source-to-source tool called ComPar on the task of detecting parallelization potential of a loop (0.8 vs 0.5 accuracy). Graph2Par, on the other hand, has shown that it can outperform PragFormer on the task of predicting OpenMP clauses applicable to a parallel loop (0.89 vs 0.85 accuracy in predicting the applicability of \texttt{private} clause). 
    More importantly, both OMPify and PragFormer have shown that they can outperform ChatGPT (GPT-3.5) on determining the parallelization potential of a loop (0.4 vs 0.86 accuracy)~\cite{kadosh2023advising}.

\end{itemize}

\subsection{Instances of HPC Enhancing LLM Performance}
HPC is crucial for efficiently enhancing LLM performance. For instance, in  Federated Learning (FL)~\cite{li2019fedavg} setups, HPC environments with heterogeneous devices can benefit from the efficient training or fine-tuning of dynamic weight-sharing models of different sizes \cite{bootstrapnas_aaai}, each selected to maximize the device's capabilities where the model is deployed. RaFFM~\cite{yu2023raffm} is an example of an FL framework that considers the resource-hungry characteristics of LLMs \cite{yu2023raffm}. This FL framework employs \emph{salient parameter prioritization} to create the abstraction of an extensive search space of efficient weight-sharing models with fewer parameters that can be efficiently extracted and deployed to an FL client with enough resources to optimize the received model while maximizing its capabilities. Low latency and high throughput are crucial for the effectiveness of this advanced FL framework. 

Sophisticated solutions must efficiently serve the optimized models in environments with diverse computing platforms. Finally, HPC challenges are also present during the LLM inference and generation stages, in which the model, either deployed at the Edge or using remote inference, must satisfy the user's efficiency and accuracy requirements. To respond to some of these challenges, the open-source community has proposed vLLM \cite{kwon2023efficient_vllm}. This library integrates with other popular libraries, e.g., Huggingface \cite{wolf2020huggingfaces}, to improve the efficiency of existing LLMs. For instance, vLLM implements the \emph{PagedAttention} \cite{kwon2023efficient_paged} algorithm, improving the management of the Key-Value cache in Transformer-based architectures, resulting in more efficient models with higher throughput. HPC can play a significant role in these solutions. 



 
\section{Future Directions and Opportunities}
\label{sec:future}
\subsection{Exploring Novel Machine Learning Problems with LLMs in HPC}
While there is much work to be done in exploring the applications of current state-of-the-art LLMs to HPC, there is ample room to explore novel problems in machine learning, such as the design of new neural network architectures and specialized learning frameworks.

The default tokenized representations of source code will be inadequate for many HPC applications. Additional input modalities that represent formal relationships in source code (e.g. control flow, data flow) as well as hardware specifications will greatly improve the utility of LLMs for HPC. To construct these architectures, we can draw inspiration from other multimodal transformers in the literature that have experimented with how different single-modality information is processed and fused \cite{nagraniAttentionBottlenecksMultimodal2021, xuMultimodalLearningTransformers2023}. These examples are predominantly focused on vision and language \cite{chaiDeepVisionMultimodal2022}, but multimodal architectures in other areas are becoming more popular as domain-specific LMs and multimodal models proliferate for various use cases \cite{krishnanSelfsupervisedLearningMedicine2022, zhengDeepScaffoldHopping2021}.

Graph representations are a particularly promising modality, as they can be accurately constructed with formal analysis techniques to represent strict relationships in code. This is in contrast to code representations learned with self-attention in LLMs, which are only probabilistic estimates of these relationships. This implies that the multimodal fusion of graph representations and LLM representations can improve the functional correctness of generated code. It seems optimal to create these graph representations by relying on message passing graph neural networks (GNNs), which only learn relationships over connected nodes, rather than reverting to graph transformers that add connections between all nodes \cite{dwivediGeneralizationTransformerNetworks2021, mullerAttendingGraphTransformers2023}. Hybrid architectures consisting of GNNs and pre-trained code LLMs would combine the best of both worlds. This would be similar to the construction of multimodal models from pre-trained vision transformers and language transformers \cite{xuMultimodalLearningTransformers2023}, but poses some additional challenges due to the inherent differences between GNNs and transformers.

Given the success of pre-trained LLMs for a variety of downstream tasks, we may hypothesize that an effectively pre-trained code LLM could be used for performance optimization problems with some fine-tuning. One intuitive way to tackle this is to fine-tune the code LLM with reward signals proportionate to the performance gain, in analogy to methods for reinforcement learning from human feedback (RLHF). In addition to LLM representations, the observation space could also contain HW properties, compiler options, and other components that would affect performance. This setup is likely to share many of the same challenges faced by research into RLHF, such as mode collapse and misgeneralization \cite{casperOpenProblemsFundamental2023}. However, this presents an opportunity to ML researchers to make further progress on these problems since HPC performance data is much more readily available than human feedback.

LLM architectures have largely been optimized for performance on GPUs, but they stand to benefit from the full use of heterogeneous HW available to machine learning practitioners. This presents the opportunity to design truly heterogeneous neural network architectures, where some components of the architecture are designed for use on GPUs, and other components are designed for use on other specialized accelerators or CPUs.


\subsection{Industry applications for Innovation in HPC with LLMs}

HPC programming requires a specialized skillset from developers that often presents a barrier to more widespread adoption. Improving the ability of LLMs to write HPC code can lower this barrier and presents an opportunity for companies to empower their developers. This may be through the creation of code-completion assistants similar to CoPilot \cite{copilot}, or through other tools that support developers, e.g. writing units, profiling performance, or fixing bugs in parallel software.

If these tools come with guarantees of correctness or scalability in specific environments, they can have a large impact in industries with demanding computational workloads. For example, biotechnology companies that invest in computational chemistry and biology would be able to extend the technical abilities of their workforce that is primarily trained in the natural sciences, rather than in computer science. Improved developer skill in HPC can also reduce the ultimate cost of cloud-based implementations of these workloads \cite{nettoHPCCloudScientific2019}.

Beyond guarantees of scalability and performance in specific environments, LLMs specialized for HPC may be particularly well-suited to S2S translation problems. Like machine translation between natural languages, this would require training on datasets in both source languages, with performance numbers for both implementations. But if they achieve success in this area, LLMs could provide solutions to cross-platform performance portability problems \cite{sathrePortabilityCPUAcceleratedApplications2019}.


A deep integration between LLMs and HPC may drive new hardware designs beyond the various specialized accelerators that have emerged for pure LLM applications. 

\subsection{Collaboration between LLM Researchers and HPC Practitioners}

As we have demonstrated throughout our paper, LLM researchers and HPC practitioners will benefit from a bidirectional exchange of expertise. Our primary goal in this work was to educate the machine learning community about key problems to solve in applying LLMs to HPC, and offer ways to apply HPC to LLMs with ever-increasing scale and complexity. While we have laid out several key opportunities at the intersection of these two fields, we are confident that ongoing research in this area will uncover new questions that will advance the state of both fields.



\section{Conclusion}
\label{sec:conclusion}
The synergy and collaboration between LLM and HPC hold the potential for mutual benefits, leading to a new era of computational efficiency. LLMs, with their advanced language processing capabilities, both encoder-decoder models and prompt-based techniques, can enhance the efficiency of HPC tasks such as code analysis, compiler optimization and transformation, data analysis, scheduling, autotuning, and storage and runtime optimization. HPC systems, known for their exceptional processing speed and parallel computing capabilities, can empower LLMs by accelerating their training processes and facilitating real-time applications. The fusion of these technologies results in a dynamic interplay where LLMs contribute to an advanced understanding of HPC applications and ecosystems. At the same time, HPC elevates the scale and speed of LLM computations, ultimately enhancing the overall performance and applicability of both technologies. This synergistic relationship has the potential to reshape the landscape of computation, opening doors to unprecedented advancements in various fields, from artificial intelligence to scientific computing.

\section*{Impact Statements}
This paper presents work whose goal is to advance the field of Machine Learning. There are many potential societal consequences of our work, none which we feel must be specifically highlighted here


\bibliography{main}
\bibliographystyle{icml2024}

\newpage
\appendix
\onecolumn
\section{HPC Downstream Tasks} 
\label{sec:appendix-tasks}

\begin{itemize}[leftmargin=*]

    \item \emph{Code-2-Code Retrieval: \cite{niu2023fair}} With a source code provided as the query, the objective of the code-to-code retrieval task is to identify codes within a pool of candidates that share similar semantics. The purpose of this task is to assess a model's capability to differentiate between codes or intermediate representations (IRs) that possess distinct semantics.
    \item \emph{Heterogeneous Device Mapping: \cite{niu2023fair} }The process of heterogeneous device mapping involves selecting the execution device that offers optimal performance for a given OpenCL Kernel, taking into consideration factors such as Input Data Size and Work Group Size (i.e., the quantity of threads collaborating in a group with shared memory).
    \item \emph{Thread Coarsening Factor Prediction: \cite{niu2023fair}} The objective here is to forecast the most efficient thread coarsening factor for a given OpenCL kernel, where this factor signifies the number of threads to combine or merge.
    \item \emph{Predicting Performant Parallel Styles: \cite{10.1145/3581784.3607038}}The objective here is to predict what kind of a parallel style is being used in the block of code in order to optimize its efficiency to its utmost limit.
    \item \emph{Triviality Detection: \cite{10.1145/3581784.3607052}}Thorough examinations reveal that simple operations are activated under specific straightforward conditions. The analysis condenses the reasons behind software trivialities into four patterns: heavy operation, trivial chain, redundant backward slice, and conditional trivial linkage. These identified patterns serve as valuable tools for detecting additional instances of trivialities and optimizing performance.
    \item \emph{Automatic Loop Fusion: \cite{10.1145/3581784.3607097}}Sparse fusion, a method designed to generate a streamlined schedule and fused code by combining a sparse kernel with loop-carried dependencies and another sparse kernel. The sparse fusion approach employs an inspector that applies an innovative iteration composition and ordering (ICO) runtime scheduling algorithm to the Directed Acyclic Graphs (DAGs) of the two input sparse kernels. ICO utilizes a vertex dispersion strategy to evenly distribute workloads in the fused schedule, incorporates two new iteration packing heuristics to enhance data locality by leveraging spatial and temporal characteristics of the merged computations, and employs vertex pairing strategies to aggregate iterations without explicitly joining the DAGs.
\end{itemize}

\end{document}